\pdfoutput=1

\documentclass[11pt]{article}

\usepackage[final]{EMNLP2023}

\usepackage{times}
\usepackage{latexsym}

\usepackage[T1]{fontenc}

\usepackage[utf8]{inputenc}

\usepackage{microtype}

\usepackage{inconsolata}
\usepackage{graphicx}

\title{Multilingual Transfer and Domain Adaptation for Low-Resource Languages of Spain}

\author{
  Yuanchang Luo, Zhanglin Wu, Daimeng Wei, Hengchao Shang, Zongyao Li,    \\
  \textbf{
    Jiaxin Guo, Zhiqiang Rao, Shaojun Li, Jinlong Yang, 
  }\\
  \textbf{
    Yuhao Xie, Jiawei Zheng Bin Wei, Hao Yang
  }\\
  Huawei Translation Service Center, Beijing, China\\
  \{luoyuanchang1,wuzhanglin2,weidaimeng,shanghengchao,lizongyao,\\
guojiaxin1,raozhiqiang,lishaojun18,yangjinlong7,xieyuhao2,\\
zhengjiawei15,weibin29,yanghao30\}@huawei.com \\
  }

\begin{document}
\maketitle
\begin{abstract}
This article introduces the submission status of the Translation into Low-Resource Languages of Spain task at (WMT 2024) by Huawei Translation Service Center (HW-TSC). We participated in three translation tasks: spanish to aragonese (es$\rightarrow$arg), spanish to aranese (es$\rightarrow$arn), and spanish to asturian (es$\rightarrow$ast). For these three translation tasks, we use training strategies such as multilingual transfer, regularized dropout, forward translation and back translation, labse denoising, transduction ensemble learning and other strategies to neural machine translation (NMT) model based on training deep transformer-big architecture. By using these enhancement strategies, our submission achieved a competitive result in the final evaluation.
\end{abstract}
\section{Introduction}
Neural machine translation (MT) \cite{lyu2019advances, bahdanau2014neural, gehring2017convolutional} allows translation systems to be trained end-to-end without having to deal with issues like word alignment, translation rules, and complex decoding algorithms that characterize statistical machine translation systems (SMT) \cite{koehn2007moses}. Recently, MT technology has evolved towards large language models (LLMs) \cite{DBLP:conf/naacl/GuoYLWSC24}. Although neural machine translation has developed rapidly in recent years, it relies heavily on big data - large-scale, high-quality bilingual corpora. Due to the cost and scarcity of real corpora, synthetic data plays an important role in improving translation quality. Existing methods for synthesizing data in NMT focus on leveraging monolingual data during training. Among them, forward translation \cite{abdulmumin2021enhanced}, back translation \cite{abdulmumin2021enhanced} and data diversity \cite{nguyen2020data} have been widely used to generate synthetic bilingual corpora. Such synthetic data can be used to improve the performance of NMT models\cite{wu2023hw}. \cite{DBLP:conf/acl/WeiWSLWGCYY23} also considers the style of the training data and exploits it to improve performance. Although synthetic data is efficient, synthetic data inevitably contains noise and erroneous translations. Denoising can prevent the training of NMT models from being interfered by noisy synthetic data by introducing high-quality real data as guidance.
Another direction to improve the performance of NMT models is to use more efficient training strategies. For example, by mixing similar language data together to train a multi-language pre-training model \cite{li2022hw}, due to the shared vocabulary, encoding layer and decoding layer parameters and language similarity, languages with less data can benefit from languages with more data. Regularized dropout \cite{wu2021r} allows the NMT model to more effectively utilize limited data during the training process. Transduction ensemble learning \cite{wang2020transductive} can aggregate the translation capabilities of multiple models into one model.


For the Translation into Low-Resource Languages of Spain task of WMT 2024, we participated in the es$\rightarrow$arg, es$\rightarrow$arn and es$\rightarrow$ast language pair. We use training strategies such as multi-language pre-training models \cite{li2022hw}, regularized dropout \cite{wu2021r}, forward translation \cite{abdulmumin2021enhanced}, back translation \cite{abdulmumin2021enhanced}, Labse denoise \cite{feng2020language} and transduction ensemble learning \cite{wang2020transductive} to train neural machine translation (NMT) models based on deep Transformer architecture.

Next, this article will expand on the details of our translation system in different translation tasks. The structure of the remaining sections is as follows: Section 2 introduces the data scale and data preprocessing process; Section 3 describes the overview of the NMT system; Section 4 gives the parameter settings, data processing results and experimental results; Section 6 gives System conclusions were drawn.
\section{Dataset}
\subsection{Data Size}
In accordance with the requirements of the WMT 2024 outline, on the Translation into Low-Resource Languages of Spain machine translation task, we used the officially provided data to train the NMT system from scratch. Table \ref{data1} shows the training data size for each language pair of the bilingual machine translation task. These language pairs include Spanish to Aragonese (es$\rightarrow$arg), Spanish to Arabic (es$\rightarrow$arn) and Spanish to Asturian (es$\rightarrow$ast).
\begin{table}[ht]
\centering
\footnotesize
\begin{tabular}{@{}lccc@{}}
\hline
& es$\rightarrow$arg & es$\rightarrow$arn & es$\rightarrow$ast \\
\hline
Bilingual & 0.06M & 2.04M & 13.36M\\
Source Monolingual & 0.4M & 8M  &8M\\
Target Monolingual & 0.26M & 6M  & 3M\\
\hline
\end{tabular}
\caption{\centering Data size for each bilingual machine translation task}\label{data1}
\end{table}
\subsection{Data Pre-processing}
The data pre-processing process is as follows:

\begin{itemize}
    \item Remove duplicate sentences or sentence pairs.
    \item Remove invisible characters and xml escape characters.
    \item Convert full-width symbols to half-width symbols.
    \item Use fast\_align \cite{dyer2013simple} to filter poorly aligned sentence pairs.
    \item Filter out sentences with more than 80 tokens in bilingual data.
    \item Remove sentences with duplicate tokens.
    \item When performing subword segmentation, joint sentencepiece \cite{kudo2018sentencepiece} is used for es$\rightarrow$arg, es$\rightarrow$arn and es$\rightarrow$ast translation tasks.
\end{itemize}

\section{NMT System}
\begin{figure}[t] 
\centering
\includegraphics[width=80mm]{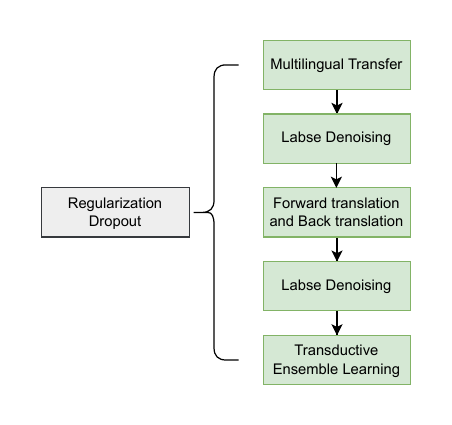}
\caption{\centering The overall training flow chart of our NMT system on the different translation tasks.}
\label{Bilingual_Training}
\end{figure}
\subsection{System Overview}

Transformer is the state-of-the-art model structure in recent MT evaluations. There are two parts of research to improve this kind: the first part uses wide networks (eg: Transformer-Big \cite{vaswani2017attention}), and the other part uses deeper language representations (eg: Deep Transformer \cite{wang2019learning}). For all MT tasks, we combine these two improvements, adopting the Deep Transformer-Big \cite{wu2023treating} model structure to train the NMT system. Deep Transformer-Big uses pre-layer normalization, features 25-layer encoder, 6-layer decoder, 16-heads self-attention, 1024-dimensional word embedding and 4096-dimensional ffn embedding.

Fig. \ref{Bilingual_Training} shows the overall training flow chart of our NMT system on the Translation into Low-Resource Languages of Spain task, we use multilingual transfer \cite{li2022hw}, regularization dropout \cite{wu2021r}, forward translation \cite{abdulmumin2021enhanced}, back translation \cite{abdulmumin2021enhanced}, Labse denoise \cite{feng2020language} and transduction ensemble learning \cite{wang2020transductive} and other training strategies are used to train neural machine translation (NMT) models based on deep Transformer-big architecture.

\subsection{Multilingual Transfer}

Recent researches have shown that multilingual models outperform their bilingual counterparts, particularly when the number of languages in the system is limited and those languages are related \cite{li2022hw}. This is mainly due to the capability of the model to learn interlingual knowledge (shared semantic representation between languages). Transfer learning using pre-trained multilingual model has shown very promising results for low resource tasks. In this task, we first select a multilingual system as the base system, then fine-tune the system with low resource language pairs.

Specifically, we add the "<arg>" tag to the Spanish side of the es$\rightarrow$arg bilingual data, the "<arn>" tag to the Spanish side of the es$\rightarrow$arn bilingual data, and the "<ast>" tag to the Spanish side of the es$\rightarrow$ast bilingual data, and sample them. Mix shuf to train a one-to-many pre-training model; sample the es$\rightarrow$arg, es$\rightarrow$arn and es$\rightarrow$ast original bilingual data and then mix shuf to train a many-to-one pre-training model.
Then, the one-to-many pre-training model and the many-to-one pre-training model are trained by using the original bilingual data, and three translation models from Spanish to Aragonese, Arabic, and Asturian and three translation models from Aragonese, Arabic, and Asturian to Spanish are obtained.

\subsection{Regularization Dropout}
Dropout \cite{srivastava2014dropout} is a widely used technique for regularizing deep neural network training, which is crucial to prevent over-fitting and improve the generalization ability of deep models. Dropout performs implicit ensemble by simply dropping a certain proportion of hidden units from the neural network during training, which may cause an unnegligible inconsistency between training and inference. Regularized Dropout (R-Drop) \cite{wu2021r} is a simple yet more effective alternative to regularize the training inconsistency induced by dropout. Concretely, in each mini-batch training, each data sample goes through the forward pass twice, and each pass is processed by a different sub model by randomly dropping out some hidden units. R-Drop forces the two distributions for the same data sample outputted by the two sub models to be consistent with each other, through minimizing the bidirectional Kullback-Leibler (KL) divergence \cite{van2014renyi} between the two distributions. In this way, the inconsistency between the training and inference stage can be alleviated.

\subsection{Forward translation and Back translation}


Forward translation, also known as self-training \cite{abdulmumin2021enhanced}, is one of the most commonly used data augmentation methods. FT has proven effective for improving NMT performance by augmenting model training with synthetic parallel data. Generally, FT is performed in three steps: (1) randomly sample a subset from the large-scale source monolingual data; (2) use a “teacher” NMT model to translate the subset data into the target language to construct the synthetic parallel data; (3) combine the synthetic and authentic parallel data to train a “student” NMT model.

Apertium is a free/open-source rule-based architecture for MT that consists of a pipeline of modules performing part-of-speech disambiguation and tagging, lexical transfer, lexical selection, chunk-level or recursive structural transfer, and morphological generation. To make our model better, we use Apertium as a "teacher" model to produce pseudo-corpus.


Back translation (BT) \cite{abdulmumin2021enhanced} refers to translating the target monolingual data into the source language, and then using the synthetic data to increase the training data size. This method has been proven effective to improve the NMT model performance.

We use the machine translation model obtained by Multilingual Transfer to produce back translation synthetic parallel data, and mix it with forward translation synthetic parallel data and authentic parallel data for training, which can achieve better results than FT or BT.

\subsection{Labse Denoising}
Due to the low quality of our bilingual data, we use LaBSE \cite{feng2020language} to calculate the semantic similarity of each bilingual sentence pair and exclude bilingual sentence pairs with similarity scores below 0.7 from our training corpus. Use these clean data to better train the model.

\subsection{Transductive Ensemble Learning}

Ensemble learning \cite{garmash2016ensemble}, which aggregates multiple diverse models for inference, is a common practice to improve the accuracy of machine learning tasks. However, it has been observed that the conventional ensemble methods only bring marginal improvement for NMT when individual models are strong or there are a large number of individual models. Transductive Ensemble Learning (TEL) \cite{wang2020transductive} study how to effectively aggregate multiple NMT models under the transductive setting where the source sentences of the test set are known. TEL uses dev sets finetune a strong model, which boosts strong individual models with significant improvement and benefits a lot from more individual models.
\begin{table*}
\centering
\begin{tabular}{lllllll}
\hline
 & \multicolumn{3}{c}{BLEU} & \multicolumn{3}{c}{ChrF++} \\
\hline
FLORES+ dev sets & es$\rightarrow$arg & es$\rightarrow$arn & es$\rightarrow$ast &  es$\rightarrow$arg & es$\rightarrow$arn & es$\rightarrow$ast\\
\hline
NMT baseline & 38.5 & 8.5 & 17.3 & 64.6 & 34.3 & 46.6 \\
+ FT \& BT & 41.7 & 9.5 & 16.9 & 64.8 & 34.9 & 45.5 \\
+ Labse denoising & 48 & 10.1 & 17.5 & 72.4 & 38.8 & 47.5 \\
\hline
FLORES+ devtest sets & es$\rightarrow$arg & es$\rightarrow$arn & es$\rightarrow$ast &  es$\rightarrow$arg & es$\rightarrow$arn & es$\rightarrow$ast\\
\hline
+ TEL  & \textbf{63} & \textbf{26.3} & \textbf{19.8} & \textbf{80.3} & \textbf{47.9} & \textbf{52.2}  \\
\hline
\end{tabular}
\caption{BLEU and ChrF++ scores of es$\rightarrow$arg, es$\rightarrow$arn and es$\rightarrow$ast NMT systems}\label{Metrics}
\end{table*}

\section{Experiment}
\subsection{Setup}
We use the open-source fairseq \cite{ott2019fairseq} to train NMT models, and then use SacreBLEU \citep{post-2018-call} and Chrf++ to measure system performance. The main parameters are as follows: each model is trained using 8 V100 GPUs, batch size is 4096, parameter update frequency is 1, and learning rate is 5e-4. The number of warmup steps is 4000, and model is saved every 1000 steps. The architecture we used is described in section 3.1. We adopt dropout, and the rate varies across different training phases. R-Drop \cite{srivastava2014dropout} is used in model training, and we set $\lambda$ to 5.

\subsection{Data processing}
\begin{table}[ht]
\begin{center}
\footnotesize
\begin{tabular}{@{}lccccc@{}}
\hline
 &&& es$\rightarrow$arg & es$\rightarrow$arn & es$\rightarrow$ast \\
\hline
Bilingual &&& 0.06M & 2.04M & 13.36M\\
Data Pre-processing &&& 0.04M & 1.51M & 3.91M\\
Labse Filter &&& 0.03M & 1.16M & 1.92M\\
Upsampling &&& 0.56M & 1.74M & 1.92M\\
\hline
\end{tabular}
\caption{\centering Data size for each bilingual machine translation task after data pre-processing}\label{data2}
\end{center}
\end{table}
Due to the poor quality of bilingual data in low-resource languages, after the rule cleaning mentioned in section 2.2 and the labse model cleaning mentioned in section 3.2, the amount of data is smaller, and the data amount of es$\rightarrow$arg, es$\rightarrow$arn and es$\rightarrow$ast is quite different. When training one-to-many and many-to-one pre-training models, if the amount of bilingual data for a certain language direction is too small, the translation quality will be extremely poor. Therefore, Following \cite{conneau2019cross, liu2020multilingual} we re-balance the training set by upsampling data from each language $l$ with a ratio: $$\lambda_l = \frac{1}{p_l}\frac{p_l^{1/T}}{\sum_{l=1}^{n} p_l^{1/T}} \quad with \quad p_l = \frac{n_l}{\sum_{l=1}^{n} n_l}$$
where, $T$ is the temperature parameter and we set $T$ to 2. $n_l$ is the number of utterances for language l in the training set. The data amount changes as shown in the following table \ref{data2}.

\subsection{Results}

Tables \ref{Metrics} shows the evaluation results of es$\rightarrow$arg, es$\rightarrow$arn and es$\rightarrow$ast NMT systems on the brand new FLORES+ dev sets and devtest sets, the results of dev test sets are obtained through OCELoT submission. We use Multilingual Transfer and R-Drop to build a strong baseline, then use FT and BT for data enhancement, and use Labse denoising for more efficient training, and finally use Transductive Ensemble Learning to ensemble multiple models ability.


As can be seen from the table above, after FT \& BT and Labse denoising, the translation quality from Spanish to three directions has been improved to varying degrees. This shows that for low-resource scenarios, these two strategies can expand the amount of data and improve the quality of the data. Enhance the translation quality of machine translation models. Among them, the improvement of both strategies in the es$\rightarrow$arg direction is higher than that of the other two directions, and the bilingual data of es$\rightarrow$arg is also the least. This shows that FT \& BT's strategy of expanding the amount of data and labse denoising's strategy of improving data quality are both in situations where the amount of bilingual data is small, The effect is more obvious. 

In addition, after Transductive Ensemble Learning, the BLEU value of FLORES+ devtest sets has been greatly improved compared to the FLORES+ dev sets test set. Although it is not the same test set, the BLEU value has improved across latitudes, which shows that  The fields of dev sets and devtest sets are very consistent, and Transductive Ensemble Learning, a strategy that utilizes dev sets, can maximize the translation effect of the model on the test set in the same field.

\section{Conclusion}
This paper presents HW-TSC's submission to the Translation into Low-Resource Languages of Spain task of WMT 2024. For both translation tasks, we use a series of training strategies to train NMT models based on the deep Transformer-big architecture. By using these enhancement strategies, our submission achieves a competitive result in the final evaluation. For example, \#607 in the spanish to aragonese constrained submissions, \#608 in the spanish to aranese constrained submissions, and \#606 in the spanish to asturian constrained submissions.

\bibliography{emnlp2023}
\bibliographystyle{acl_natbib}

\end{document}